\documentclass{article}

\usepackage{arxiv}
\usepackage{graphicx}

\usepackage{cite}

\graphicspath{{images/}{../images/}}
\usepackage{pdflscape}
\usepackage{algorithm}
\usepackage[noend]{algpseudocode}

\usepackage{array}

\usepackage{comment}

\usepackage[lofdepth]{subfig}

\usepackage{multirow}

\usepackage[utf8]{inputenc}

\title{Cell Nuclei Classification In Histopathological Images using Hybrid O\textsubscript{L}ConvNet}

\author{
 Suvidha Tripathi \\
  Department of Information Technology\\
  Indian Institute of Information Technology Allahabad\\
  Jhalwa, Deoghat, Prayagraj, Uttar Pradesh 211015 \\
  \texttt{suvitri24@gmail.com} \\
   \And
 Satish Kumar Singh \\
  Department of Information Technology\\
  Indian Institute of Information Technology Allahabad\\
  Jhalwa, Deoghat, Prayagraj, Uttar Pradesh 211015 \\
  \texttt{sk.singh@iiita.ac.in} \\
}

\begin{document}
\maketitle
\begin{abstract}
Computer-aided histopathological image analysis for cancer detection is a major research challenge in the medical domain. Automatic detection and classification of nuclei for cancer diagnosis impose a lot of challenges in developing state of the art algorithms due to the heterogeneity of cell nuclei and data set variability. Recently, a multitude of classification algorithms has used complex deep learning models for their dataset. However, most of these methods are rigid and their architectural arrangement suffers from inflexibility and non-interpretability. In this research article, we have proposed a hybrid and flexible deep learning architecture O\textsubscript{L}ConvNet that integrates the interpretability of traditional object-level features and generalization of deep learning features by using a shallower Convolutional Neural Network (CNN) named as $CNN_{3L}$. $CNN_{3L}$ reduces the training time by training fewer parameters and hence eliminating space constraints imposed by deeper algorithms. We used F1-score and multiclass Area Under the Curve (AUC) performance parameters to compare the results. To further strengthen the viability of our architectural approach, we tested our proposed methodology with state of the art deep learning architectures AlexNet, VGG16, VGG19, ResNet50, InceptionV3, and DenseNet121 as backbone networks. After a comprehensive analysis of classification results from all four architectures, we observed that our proposed model works well and perform better than contemporary complex algorithms.
\end{abstract}

\keywords{Deep Learning, Hybrid networks, Object level features, Transfer Learning, Histopathological Images, Cell Nuclei Classification, Class balancing, Convolutional Neural Networks, Multi Layer Perceptron}

\section{Introduction}
\par Early cancer detection is a major challenge in the medical domain. Even today the medical community is largely dependent upon the expert pathologist for detecting and classifying such cell anomalies that  cause cancer, in whole slide histopathological images. The job of the pathologists becomes very cumbersome and may take several days for annotating the whole slide images of biopsy samples. Moreover, the reliability of predictions also depends upon the experience of the pathologist and some times, consensus of more than one pathologists are required for confirming such anomalies. These factors provide adequate motivation for research and development of a computer-assisted diagnostic (CAD) systems which classifies cell nuclei and improves the understanding of some of the underlying biological phenomenon, e.g., monitoring cancer cell cycle progress \cite{chen2006automated}, type, shape, size and arrangement of the cells in the affected organ sites, and the knowledge about metastasis, if the cells are present at some unlikely locations . All these observations can be comprehended if we know the type of cell present in the diseased tissue sample. Early diagnosis of cell anomalies can largely affect the disease prognosis \cite{tumour}. Such as in the case of a colon or colorectal carcinoma, epithelial cells lining the colon or rectum of the gastrointestinal tract are affected and timely detection of these cells can help in quick diagnosis, which eventually would increase the prognostic value of the disease. Similarly, the lymphocytes can also be analyzed for sentinel lymph node disease \cite{tumour}. The other examples are the Myeloma or multiple Myeloma detections through the plasma cells which are the types of the white blood cells and cause the cancer \cite{cancer17}. Therefore, a sample biopsy from a specific location can be quickly analyzed using the information of cell environment provided by appropriate CAD system. 

\par In particular medical image analysis, for all diagnosis, is attributed to the knowledge and skills possessed by trained and experienced pathologists. Although pathologists have the ability and means to single out the affected cancerous lesions in a tissue biopsy samples, most of such detections are still done manually and hence time-consuming. Numerous challenges are involved in diagnosing cancer due to data set variability and heterogeneity in cell structures, which makes the process extremely tedious even for experts. Software intervention for early detection is therefore important for the purpose of effective control and treatment of the diseased organs \cite{madabhushi2016image}. To develop such automated cell detection and classification algorithms, the knowledge of histology is vital and requires the annotated or labeled data set to be prepared by the expert histo-pathologists. Once the labelled data is acquired, then the routine intervention of pathologists can be eliminated while analyzing the whole slide samples under test by using the developed automated CAD algorithms. Cell nuclei in a Hematoxylin and Eosin (H\&E) stained histopathological slide sample have a specific shade of blue caused by hematoxylin's reaction with cellular protein present in the nuclei of the cells \cite{fischer2008hematoxylin}. A shape of the cell varies with cell type, cell-cycle stage, and also with the presence or absence of cancer. Fig. 1 shows four different classes of nuclei namely, Inflammatory, Fibroblast, Epithelial, and Miscellaneous, which include adipocyte, endothelial nucleus, mitotic figures, and necrotic nucleus \cite{7848388}. The nuclei structures as shown in Fig.1 have different shapes, texture, and intensity features, which vary by the factors i.e. nuclei type (epithelial, fibroblasts, lymphocytes, etc.), the malignancy of the disease (or grade of cancer), and nuclei life cycle (interphase or mitotic phase) \cite{zink2004nuclear}. 
\begin{figure}
\centering
\includegraphics[width=3in]{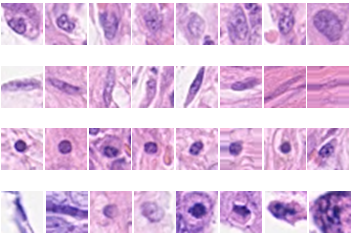}
\caption{ Example sub-images of different classes of nuclei starting from first row to fourth:  Epithelial, Fibroblasts, Inflammatory, Miscellaneous (in sets of 2 - Adipocyte, Endothelial, Mitotic Figure, and Necrotic Nucleus (from left to right) }
\label{fig1}
\end{figure}
For example, the inflammatory nuclei, a type of white blood cells and called as lymphocyte nuclei (LN) are smaller in size and regular spherical shape as compared to epithelial nuclei (EN) \cite{irshad2014methods}. Fibroblasts have long spindle-like shape and appearance and having very little cytoplasm \cite{rodemann2011functional}. Activated fibroblasts and infiltrating inflammatory cells are the signs of potential tumor growth \cite{rodemann2011functional}. All these histological and biological differences between cell structures and their site of origin highlight the clinical relevance of classifying different types of nuclei.

\par In this paper, we have undertaken the feature based approach for automated nuclei classification. Feature based approach can be classified into two general categories: hand-crafted features and deep learning based features. In histopathology Images, morphological and architectural features whose accuracy depends on the amount of magnification and the type of class and which has unique mixture of visual pattern qualify as hand crafted features whereas unsupervised deep learning features are intuitive and are a by-product of filter responses obtained from large number of training samples and fine tuning of the network. In the proposed work we clearly proved the benefits of using combined feature set consisting both object level features and learned deep learning features over feature set acquired from single domain on complex medical data. Moreover, for detailed analysis, the accuracy-generalization tradeoff and space-time complexity issues as exhibited by traditional and DL methods respectively have been considered in the proposed architectural arrangement. 
 
In summary, key contributions of this work include:
\begin{enumerate}
\item The strength of our method lies in the flexible architecture that supports a backbone of deep learning model to extract deep features and a simple object level extraction framework for extracting cell level features.
\item We achieved a high level of nuclei classification system through simple concatenation of derived features from two domains. 
\item The emphasis has been put through a series of experiments that in case of nuclei structures, even very small number of basic and locally focussed object level features can enhance the performance if combined with three or more layers of deep learning architecture. 
\item To the best of our knowledge, this is the first study on this hypothesis that have developed a custom architecture for the problem to highlight the need of designing lighter architectures for specific problems rather than using deeper pre-trained architectures.
\item To the best of our knowledge, this is the first study on this hypothesis to have experimentally prove the performance of end to end learning over stage wise learning. 
\end{enumerate}

\par The rest of the paper is organized as follows. Section-II describes the reviewed literature for handcrafted and deep features. Section-III describes the complete methodology of the proposed work. The experimental setup is elaborated in Section-IV including the database and workflow. Section-V contains various results and necessary discussion. Discussion section also justifies the appropriateness of the proposed method while the flexibility and robustness are the points of concerns. Section-VI concludes the work presented in the paper followed by the acknowledgments and references.

\section{Reviewed Literature}
\par Owing to the above-mentioned properties exhibited by cell nuclei, many traditional handcrafted cell nuclei classification algorithms have been reported in \cite{wang2016automatic, diamond2004use, doyle2007automated, gurcan2009histopathological, boucheron2008object, sirinukunwattana2015novel, shi2017histopathological}. Authors in \cite{wang2016automatic} have first segmented the nuclei objects using morphological region growing, wavelet decomposition and then found the shape and texture features for classifying cancer cells vs. normal cells using SVM classifier. Another handcrafted feature-based method for cell nuclei classification in histopathological images while using shape, statistical, and texture (Gabor and Markov Random Field) features from localized cells has been reported in \cite{diamond2004use}. 
Other object level (OL) feature extraction from localized cell objects based methods have been reported in \cite{doyle2007automated}. All the methods \cite{wang2016automatic, diamond2004use, doyle2007automated, sirinukunwattana2015novel, shi2017histopathological}  using the OL features have been critically analyzed against the utility of those features for individual problems related to a histological and/or cytological analysis of cancer cells in \cite{gurcan2009histopathological,boucheron2008object}. The quality of the extracted features from various handcrafted methods \cite{wang2016automatic, diamond2004use, doyle2007automated, gurcan2009histopathological, boucheron2008object, sirinukunwattana2015novel, shi2017histopathological} is then assessed after passing them through appropriate classifiers. The success of their findings motivated the use of targeted OL features in our methodology. To design effective handcrafted feature-based models requires complex algorithms to achieve high performance and a decent level of domain-specific knowledge \cite{sirinukunwattana2015novel, shi2017histopathological}. Moreover, it becomes extremely difficult to resolve the issues due to dataset variability and heterogeneity within each cell type. These issues lead to the inability of the reported novel but complex models to generalize well with varying datasets. It is worth mentioning that, most of these methods are reported on a very small sample size in general, causing robustness issues. To overcome the generalization problem, it is required to model features, which are common for a particular class of cell nuclei but highly discriminating among different classes. 
Recently, deep learning architectures have been known to produce generalized feature sets and hence have proved their niche in classification algorithms \cite{lecun2015deep, szegedy2015going, he2016deep, simonyan2014very, bejnordi2017diagnostic, zhang2018classification, wang2014mitosis, cirecsan2013mitosis}. To put it more clearly, the key advantage of using deep learning architectures can be explained by highlighting the problems of linear or other shallow classifiers. The traditional classifiers do not use raw pixel data and possibly cannot distinguish two similar objects on different backgrounds which is a case of selectivity-invariance dilemma \cite{lecun2015deep}. That is why we need good feature extractors with such classifiers to solve the selectivity-invariance dilemma. The Deep Learning (DL) based architectures automatically learn the good feature sets from the large histopathological image data sets. In 2016, the CAMELYON challenge has also reported the use of extensive deep learning architectures for solving various problems of Localization and Classification. Detailed methodologies of all these methods have been reported in \cite{bejnordi2017diagnostic}. More recently, authors in \cite{sajjad2019multi} have used pre-trained VGG19 to classify extensively augmented multi grade brain tumour samples whereas authors in \cite{Wang2018} to identify alcoholism in subjects using their brain scans. So, DL methods find applicability in wide range of applications due to their robust and better performing architectures. However, there are some issues with deep learning based methods as well. DL features lack interpretability and cannot be confirmed as global or local features. Moreover, there is always the lack of a large number of datasets in the medical domain, which hampers or restricts DL algorithms to scale well on all other test data sets not used for the training. Another major issue with deep architectures is the huge parameters with greater depths, causing the optimization problem very time-consuming. At the same time, the complexity of a model increases, as the depth increases and eventually the intermediate processes become less and less interpretable. One of the approaches for minimizing the training time on the medical dataset is to use the concept of transfer learning and fine-tuning pre-trained models such as  AlexNet \cite{szegedy2015going}, VGG16, VGG19 \cite{simonyan2014very}, ResNet50 \cite{he2016deep}, DenseNet \cite{huang2017densely}, and InceptionV3 \cite{szegedy2015rethinking}. Originally these models have been trained on natural image datasets, which is from an entirely different domain but can be fine-tuned to extract features on medical images. But, medical data has very little to no correspondence with natural images. Hence, relying solely on transfer learning based fine-tune approach should not be preferred. Rather, the training should be done on the networks that have either not been pre-trained on natural images or have been pre-trained on similar medical dataset. But, training DL networks on medical dataset has its own set of challenges, including lack of huge amount of annotated medical data for training. Moreover, the diverse nature of medical images prevents the generalization and standardization of datasets on which DL networks could be trained for transfer learning. 
\par An exhaustive survey of deep learning methods as reported in \cite{ravi2017deep} thoroughly highlights the merits of applying DL methods in the field of medical imaging, medical informatics, translational bioinformatics, and public health. The amalgamated use of both OL and DL features for the purpose of nuclei detection, segmentation, and classification have also been suggested in \cite{irshad2014methods,ravi2017deep}.   

\par Therefore, OL features in combination with DL features could help to bridge the gap between issues that two domains bring individually. Some recent articles have worked on the similar hypothesis of inter-domain feature combination and developed a method that combines the two feature sets as reported in \cite{zhang2018classification,wang2014mitosis}. But, the drawback of their method is in their complexity and huge training times due to very deep network models. Authors in \cite{zhang2018classification} combined different deep learning features extracted from Caffe-ref \cite{jia2014caffe}, VGG-f \cite{chatfield2014return} and VGG19 models with Bag of Features (BoF) and Local Binary Pattern (LBP) features. They then used ensemble classifiers to produce better classification accuracy than that of the softmax classification method used by deep learning models. However, the dataset under experiments in \cite{zhang2018classification} was imbalanced hence, the reported accuracy trend may not hold good for other imbalanced datasets which are highly probable in case of medical image datasets. F1 score and AUC are better parameters for assessing the performance of classification algorithms for imbalanced datasets. Also, the authors  \cite{zhang2018classification,wang2014mitosis} reported the complex models which were based on pre-trained deep architectures with 7 and more layers and did not analyze the performance trend on other customized architectures that could have minimized the space and time constraints. It is difficult to design and test such relatively inflexible algorithms on a new dataset and deploy in real time applications. For example, it is difficult to change the design if one wishes to add a new functionality and re-train the algorithm. Furthermore, the reported handcrafted features in these studies lack direct relevance to the nuclei structural properties. 

\section{METHODOLOGY}
\begin{figure*}
\centering
\includegraphics[width=\textwidth ,height=4in]{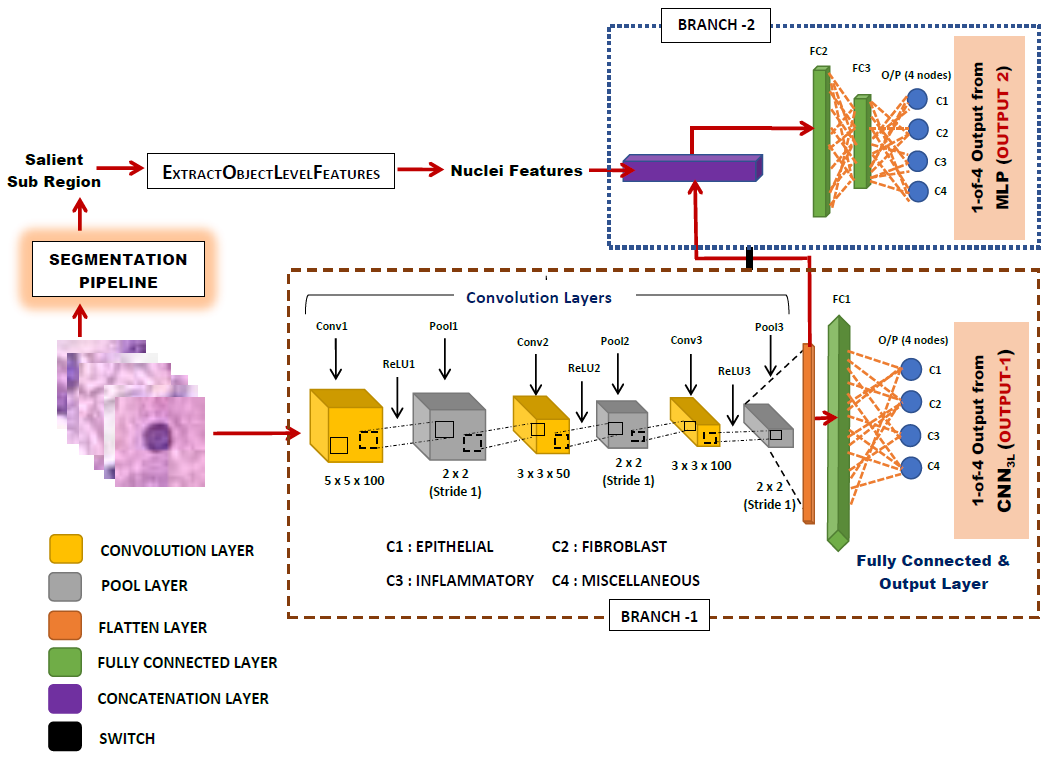}
\caption{Block Diagram of proposed OLConvNet. Raw training images of cell nuclei is passed through Branch-1 of the network for DL feature extraction and further for classification using fully connected (FC1) and softmax layer of the DL network (OUTPUT-1). OL features are extracted from segmented nuclei images after segmentation pipeline. Extracted OL features are classified in Branch 2. Switch between branch 1 and 2 make the decision about which kind of output we would want for our dataset (OUTPUT-1 or OUTPUT-2 or Both).}
\label{fig2}
\end{figure*}
Hybrid feature based flexible classification framework trained on a dataset from \cite{sirinukunwattana2016locality} is used to determine the suitability of combining different feature sets. Few pre-processing steps are performed to segment the cell nuclei from background stroma. This step is necessary to extract the OL features. This feature set comprises relevant visual, shape and texture features from each nucleus. DL feature is extracted from the original input images. Both the set of features are then fused to produce a final feature set. The final fused feature set has been used by Multi-Layer Perceptron (MLP) as input for classifying the cell nuclei into one of the four categories. The block diagram of the proposed architectural setup has been shown in Fig. \ref{fig2}.

The entire flow has been modeled in the Algorithm-1.
\begin{algorithm}
\caption{O\textsubscript{L}ConvNet}
\begin{algorithmic}
\State
\textbf{Input:}
Training data set $D_{tr}$ with \textbf{m} samples. Data $f_{i}(X)$, where i=$1,\dot{...} , m$ is an instance in the 3 dimensional image space $X \in \Re^{U \times V \times d} $ and $y_{i} \in Y = \{ 1,2,3,4\}$ is the class identity label associated with $f_{i}(X)$. 
\State 
\textbf{Output:} Output1: $\hat{y}^{cnn}$ 1-of-4 Softmax output from $CNN_{3L}$, Output2: $\hat{y}^{mlp}$ 1-of-4 Softmax output from MLP
\State \textbf{Pre - Processing (Balancing Dataset)}:
$\{f_{i}(X), y_{i}\}_{M} \gets ADASYN(\{f_{i}(X), y_{i}\}_{m}$) \Comment{\small\textit{M - samples after balancing dataset, ADASYN - a function used for balancing the dataset}}  \par
\For{$x_i \in X$} 
 		\State \textsc{Segmentation}: $x_{i}^s \gets SegN(x_{i})$  \Comment{\small\textit{SegN: a function following segmentation pipeline}} 
 		\State \textsc{ObjectLevelFeatures}: \,$(FV_{OL}^{i})_{M\times N_{1}} \gets OLFV(x_{i}^s)$ \par \Comment{\small\textit{OLFV: a function extracting nuclei features, $N_1$ - Number of Object Level features}} 	
 		\EndFor
\State\textbf{STAGEWISE - TRAINING}: 
\For {epoch $e \in (1,100)$} 
	 \For {batch b} 
	 	 \For {$x_{i} \in b$} 
	 	  \State CNN Inference: $\hat{y_{i}}^{cnn} \gets CNN_{\omega^{cnn}}(x_{i})$ 
	 	  \State $\frac{\partial{L}}{\partial{\omega^{cnn}}} \gets  \hat{y}^{cnn}_{i}, y_{i}$ \Comment{\textit{$\omega$ - weight parameter}} 
	 	  \State Compute $\frac{\partial{L}}{\partial{\omega^{cnn}}}$ using backpropagation \Comment{\textit{L - cross-entropy Loss function}} 
	 	  \State Update CNN: $\omega^{cnn} \gets \omega^{cnn} - \eta{\frac{\partial{L}}{\partial{\omega^{cnn}}}} $ \Comment{\textit{$\eta$ - Learning rate}}
	 	 \EndFor
	 \EndFor
\EndFor
\State ExtractFeatures: $\{FV_{cnn}\}_{(M \times N_{2})} \gets CNN_{3L}(\{f(X)\}, layer)$ \Comment{\textit{$N_2$ - Number of CNN features}}
\State Concatenation: $\{TotalFeatures\}_{M \times(N_{1} + N_{2})} \gets FV_{cnn} + FV_{OL}$ 	 	  
\For {$TF_{i} \in TotalFeatures$} 
	\State MLP Inference: $\hat{y_{i}}^{mlp} \gets MLP_{\omega^{mlp}}(TF_{i})$
	\State $\frac{\partial{L}}{\partial{\omega^{mlp}}} \gets  \hat{y}^{mlp}_{i}, y_{i}$
	\State Compute $\frac{\partial{L}}{\partial{\omega^{mlp}}}$ using backpropagation
	\State Update MLP: $\omega^{mlp} \gets \omega^{mlp} - \eta{\frac{\partial{L}}{\partial{\omega^{mlp}}}}$
\EndFor 
\textbf{END to END - TRAINING}:\par
\For {epoch $e \in (1,100)$} 
 	 \For {batch b}
         \For{ $x_{i} \in b$ }
	 	 \State CNN Inference: $\hat{y_{i}}^{cnn} \gets CNN_{\omega^{cnn}}(x_{i})$ 
	 	  \State $\frac{\partial{L}}{\partial{\omega^{cnn}}} \gets  {\hat{y}^{cnn}_{i}}, y_{i}$ 
	 	  \State Compute $\frac{\partial{L}}{\partial{\omega^{cnn}}}$ using backpropagation 
	 	  \State Update CNN: $\omega^{cnn} \gets \omega^{cnn} - \eta{\frac{\partial{L}}{\partial{\omega^{cnn}}}} $ 
	 	  \State ExtractFeat: $\{FV_{cnn}^i\}_{(1\times N_{2})} \gets CNN_{3L}(\{f_{i}(X)\},layer)$
	 	  \State Concatenation: $\{TotalFeatures^i\}_{1\times (N_{1} + N_{2})} \gets FV_{cnn}^i + FV_{OL}^i$ 	 	
	 	  \State MLP Inference: $\hat{y_{i}}^{mlp} \gets MLP_{\omega^{mlp}}(TotalFeatures^{i})$ 
	 \State $\frac{\partial{L}}{\partial{\omega^{mlp}}} \gets  \hat{y}^{mlp}_{i}, y_{i}$ 
	 \State Compute $\frac{\partial{L}}{\partial{\omega^{mlp}}}$ using backpropagation 
	 \State Update MLP: $\omega^{mlp} \gets \omega^{mlp} - \eta{\frac{\partial{L}}{\partial{\omega^{mlp}}} }	$ 
	 \EndFor
	 \EndFor
	 \EndFor
\end{algorithmic}
\end{algorithm}

Various steps involved in the proposed methodology has been elaborated in the following sub-sections (3.1)-(3.5)

\subsection{Segmentation}
\par Cytologic and histologic images prevent the generalization of segmentation algorithms because of the inherent variability of the nuclei structures present in them. Due to this reason, determining which state of the art algorithm for nuclei segmentation would work for our dataset was a lengthy problem. Therefore, we seek to develop application-specific segmentation algorithm for OL feature extraction.  Our dataset contains H\&E (Hematoxylin and Eosin) stained RGB image blocks that stains the nuclei region in bright blue and cell region in pink. The staining helped us to roughly extract the nucleus contour. Segmentation of an object then allowed for the calculation of OL features such as homogeneous color, texture, size, and shape of the segmented region. 

\par
\begin{figure}[htbp]
\centering
\includegraphics[width=\textwidth ,height=2in]{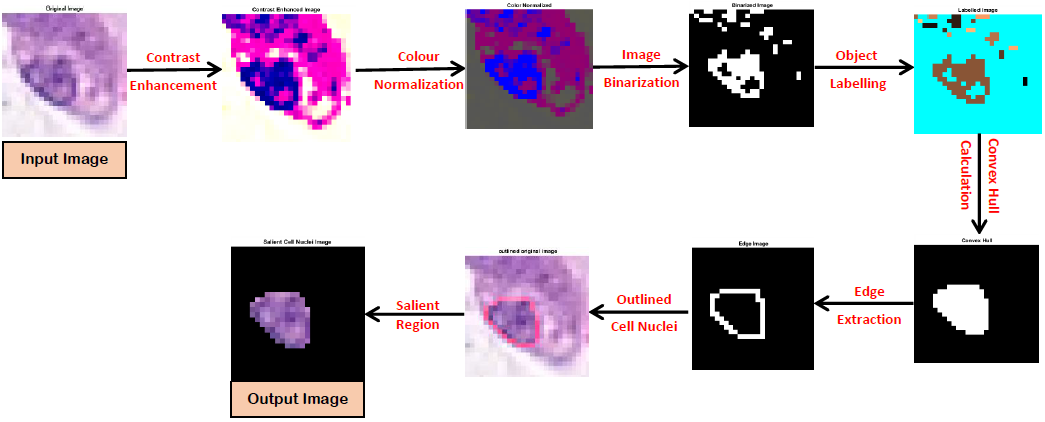}
\caption{Segmentation Pipeline of our network}
\label{seg}
\end{figure}

\begin{figure}[htbp]
\centering
\includegraphics[width=3.5in ,height=2in]{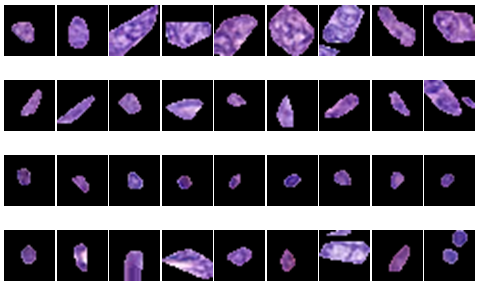}
\caption{ Example segmented-images of different classes of nuclei starting from first row to fourth:  Epithelial nuclei, Fibroblasts, Inflammatory nuclei, and Miscellaneous }
\label{segex}
\end{figure}

 Firstly, we enhanced the blue intensity of the nuclei through contrast adjustment. For this purpose blue color channel intensities were mapped from initial values to 255. Similarly, Red and Green channel pixel values less than a certain range were also tweaked towards higher range. This technique of adjusting intensity values in each channel to new values in an output image helped in highlighting poorly contrasted nuclei regions from cell cytoplasm and background noise. We assigned a higher value to blue intensity pixels relative to red and green components because blue-ratio is proven to be capable of highlighting nuclei regions in H\&E stained histopathological images \cite{chang2012nuclear}. This step is followed by color normalization so that the intensity values follow normal distribution and as well remove any noise/artefact that may have been introduced due to contrast enhancement. For the next step, we computed the binary image and calculated the convex hull of the labelled region having the highest number of pixels. Convex hull of the binary image ensured that the largest area containing most blue pixels is retained and the defined boundary of the nuclei can be obtained for calculating OL features. In other words, the perturbation in nuclei structure due to staining process may distort original nuclei structures, so obtaining a convex hull defines the smooth boundary around the nucleus. This further helps in following procedural steps of extracting OL features.   Convex hull step is then followed by edge extraction of the convex hull. Lastly, we did the scalar multiplication of the resultant image with the original image to obtain the final output of a segmented RGB Nuclear image. Segmentation results helped in delineating nuclei region from the surrounding tissues. Figure \ref{seg} shows the pipeline of segmentation.  Some of the segmented classwise nuclei examples are shown in Figure \ref{segex}. 


\subsection{Object Level Feature Extraction}

\par We have extracted a set of nine features. These nine features include color, texture and shape information of the nuclei calculated on the segmented nuclei image.  Color information contains the pixel intensity value that has the highest frequency in the histogram. Since the nuclei of the cells take the shades of blue after H\&E staining, the area which has the maximum intensity of blue would be the area inside the nucleus. If the intensity with maximum frequency is not around the blue range, it is not the nucleus and that acts as the differencing factor while classifying nuclei against miscellaneous or poorly segmented region. Texture information contains the GLCM texture features \cite{haralick1973textural}. 
Among many GLCM texture features, we calculated four statistical texture features which are Contrast, Homogeneity, Correlation, and Energy of the nucleus surface. Texture, being a fundamental property of tissue surfaces, helps to differentiate between different types of cells such as epithelial, fibroblasts and lymph node cells.  In papers \cite{rodemann2011functional}, \cite{diamond2004use} and \cite{doyle2007automated} authors described the shape and morphology features of different classes of nuclei. The considered shape and morphological features in this paper are the areas, major axis length, minor axis length, and eccentricity.

\subsection{Convolutional Feature Extraction}

\par CNN Architecture: The general CNN architecture first proposed by Yann LeCun et al. in their paper \cite{lecun1989backpropagation}. In CNN, unlike traditional machine learning approaches that take features as input to classify the data into categories, raw images are used as input. It works on the input images by learning filter weights and modify itself until convergence. 
The basic architecture of CNN comprises seven major layers, namely, Input image layer, Convolution layer, ReLU layer (Non-Linearity), Pooling (Local Max), Fully connected layer, Softmax Layer, and Classification Layer.

\par An exhaustive theoretical account of CNN can be found in \cite{karpathy2018stanford}. These layers, when combined in a pattern, create deeper networks with the increased number of hidden units. Deeper the network, a more exhaustive set of features could be extracted from the image \cite{donahue2014decaf}. However, this theory is highly subjective and depends majorly on the properties of the datasets. For example, the size of the image data should be large enough to be processed into a meaningful representation in case of architectures with great depths. Also, in a few cases, the number of parameters resulting due to large depths pose a major disadvantage in terms of computational load and efficiency of the algorithm. Hence, we did experiments to develop a custom three layer convolutional network to extract features from the nuclei dataset. We named it CNN\textsubscript{3L} for ease of reference. Figure \ref{fig2} shows the network in detail.  After training the whole network for a set number of epochs, the final network yields the best set of features which are then further progressed to the classification layer. For the purpose of extracting DL features, it is preferred to extract features from the last layer located before the fully connected layer since the final convolution layer features are a more specific representation of the dataset images. Our proposed setup also has the flexibility to change the backbone architecture from shallow three-layer network to any number of layers.  

\subsection{Fusion}

\begin{figure}[htbp]
\centering
\includegraphics[height=2in]{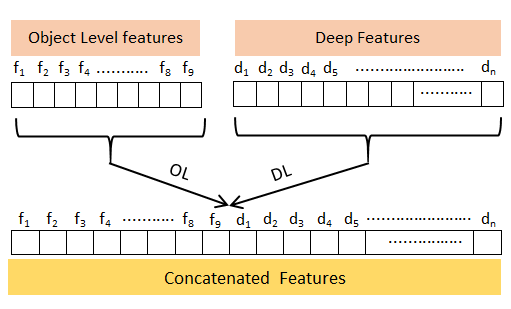}
\caption{Diagram displaying fusion of OL and DL features}
\label{fig-fusion}
\end{figure}

We get two set of features which are globally extracted from segmented nuclei (OL features) and automatically extracted features that include both local and global information (DL features) of the image. We have concatenated these two sets along the axis that contained the feature vector of a sample and produced one combined set for further classification (Fig. \ref{fig-fusion}). This exhaustive set of features are then used as input to the first MLP layer for the purpose of categorizing them into four classes. 

\subsection{Classification}

\par For classification, we have used Multi-Layer Perceptron (MLP). This process is called Transfer Learning where CNN is used only for feature extraction while the next step of classification is performed by another machine learning algorithm such as MLP for multiclass classification. In this paper, we used the MLP network with one input layer, one hidden layer having ten nodes, and one output layer. Combination of features is fed as an input to the first MLP layer. Hidden layer used tansig as an activation function defined by 
\begin{equation}
tansig(n)=\frac{2}{(1+exp(-2*n))-1}, 
\end{equation}
where $n$ is our input vector to the second hidden layer. This activation function is faster than $tanh$, however numerical differences are small \cite{vogl1988accelerating}. The output prediction scores from the output layer of MLP was given by softmax function \cite{bishop2006pattern} -\cite{murphy2012machine} defined as: 
\begin{equation}
p_{j}(\hat{y}(x))= \frac{exp(\hat{y}_{j} (x))}{\sum_{k} exp(\hat{y}_k(x))},
\end{equation}
where $\hat{y}_{j} (x)$ denotes the $jth$ element of output vector $\hat{y}(x)$
We performed 2-fold cross-validation test on our network to observe the efficacy of our model. 

\section{Experimental Setup}
\subsection{Database}
\par For our experiments, we have taken the database from \cite{sirinukunwattana2016locality}. This database has a total of 100 H\&E stained histology images of size 500 by 500 pixels. The center coordinate location of each nucleus in each image has been provided as ground truth. Different types of nuclei are divided into sets belonging to four classes which are Epithelial nuclei, Fibroblast nuclei, Inflammatory nuclei and the rest of the small types are categorized as one class called 'miscellaneous'. We obtain subimages of size 27x27 extracted around the center coordinate locations and maintained them in four sub-folders segregated by class types. Four classes in the dataset have 7,722 Epithelial, 5,712 Fibroblasts, 6,970 Inflammatory, and 2039 miscellaneous nuclei totaling up to 22,444 nuclei in 100 H\&E stained histopathological images. The number of samples in each class are imbalanced which can cause classification results biased towards the majority class ( the class having the highest number of samples). A systematic study of the class imbalance problem in convolutional neural networks by authors in \cite{buda2018systematic} have concluded through their research that the class imbalance problem, if not addressed may have a detrimental effect on classification performance. The influence of imbalance on classification performance increases with the size of the dataset. They also reported that in case of convolutional neural networks, the method that works best for eliminating class imbalance problem is oversampling with thresholding which opposed to other machine learning models, does not cause overfitting in CNN. Thus, we performed our experiments after balancing our dataset. 
We used adaptive synthetic sampling for eliminating class imbalance by synthetically creating new examples in the vicinity of the boundary between the two classes than in the interior of the minority class via linear interpolation between existing minority class examples \cite{he2008adasyn}. So, after creating synthetic examples for minority classes (Fibroblast, Inflammatory, and miscellaneous) with respect to number of samples in majority class (Epithelial) we accumulated 29,771 total data points having 7,722, 7,275, 6,970, and 7804 samples in class 1 (Epithelial), class 2 (Fibroblast), class 3 (Inflammatory), and class 4 (Miscellaneous), respectively. After acquiring the balanced nuclei dataset, for the purpose of training, validation, and testing, we divided each class in the ratio 0.7:0.15:0.15, respectively. All the networks are evaluated on the testing set and performance metrics are reflected in Section V.

\subsection{CNN\textsubscript{3L} and Parameter Settings}
\par
\begin{figure*}[htbp] 
\centering
 \subfloat[short for lof][Accuracy at learning rate 1e-4]{
   \includegraphics[width=0.455\textwidth, height=2in]{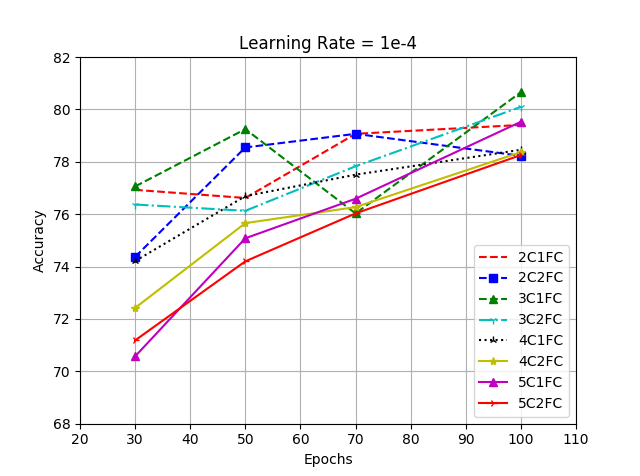}
   \label{subfig:figa}
 }
 \subfloat[short for lof][Accuracy at learning rate 1e-5]{
   \includegraphics[width=0.45\textwidth, height=2in]{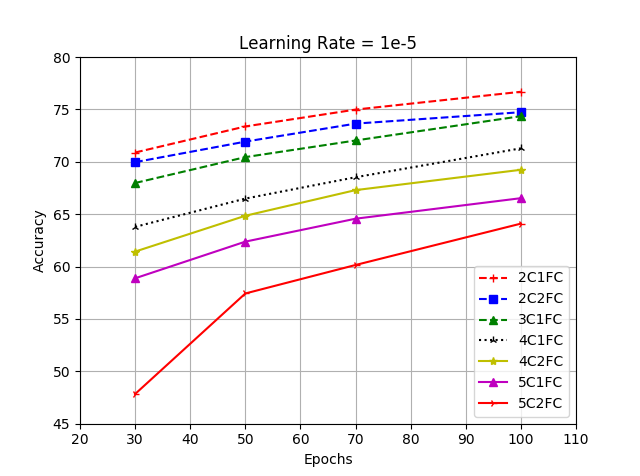}
   \label{subfig:figb}
}

\caption[short for lof]{Epochs v/s Accuracy graph at different hyperparameters}
\label{fig10}
\end{figure*}

 We used the CNN framework with three convolutional layers and one fully connected layer as shown in  Figure \ref{fig2}.  Traditionally, CONV\_RELU layers are stacked, which are then followed by POOL layer. This pattern is repeated until we get a relatively small number of parameters from the image or when the optimal performance is achieved.  The aim to stack up layers is to extract a more exhaustive set of features. 
The last fully-connected layer holds the output, such as the class scores. 
In our research work, experiments with a different number of layers yielded the optimal value of three convolution layers and one fully connected layer. We have used our knowledge about the problem, applied a heuristic approach to select parameters, and observed the outputs. Based on the outputs obtained, we tweaked the parameters. This process was repeated until we got the optimal set. The architecture was trained with different hyperparameters such as the number of convolutional layers and fully connected layers, learning rate and the number of epochs. The accuracy achieved at each instance was recorded and plotted as shown in figures \ref{subfig:figa} and \ref{subfig:figb}. The plots show the accuracy values on the Y-axis corresponding to the number of epochs on X-axis for each type of architecture taken into consideration. Here, \textit{\textbf{N}C\textbf{M}FC} denotes N Convolutional layers and M Fully Connected layers. The two figures show observations recorded from two learning rates.

\par From the experiments, we found out that neither stacking up more layers nor decreasing the learning rate showed a positive impact on the performance of the network. A learning rate of $1e-4$ with three convolutional layers and 1 fully connected layer the accuracy recorded at epoch number 100 was the highest (refer figure \ref{subfig:figa}). Increasing the learning rate further to $1e-3$ for CNN\textsubscript{3L} decreased the accuracy to 61.77\% at epoch 70 and further down to 59.41\% at epoch 100. \\
For training the final optimal network (CNN\textsubscript{3L}), we normalized our dataset images by subtracting the mean image of the training set from all training set images before using them as input vectors. If we don't scale our input training vectors, the feature values distribution range would likely be different for each feature, and thus the corrections in each dimension after each epoch through learning rate will differ for each dimension. This might cause overcompensation (very high variance from mean weight) in weight of one dimension while undercompensating (very low variance from mean weight) the another. This creates a non-ideal situation since it might get difficult to center the weight space.  This will also affect the time efficiency as it will travel too slow to get to a better maxima. 

\par In our architecture CNN\textsubscript{3L}, We followed 5 x 5, 100 channel, 3 x 3, 50 channels, and 3 x 3, 100 channels for each convolutional layer respectively. We experimented with different filter window size and numbers of filters and observed little to no improvement in the performance, for example, experiment with filter numbers 32, 64 and 128 on subsequent three convolutional layers of the optimized CNN\textsubscript{3L} achieved the F1-score of 0.7907. There could be an infinite number of permutations to choose network parameters from. We performed numerous experiments to decide on CNN\textsubscript{3L} as our optimal network, however finding a theoretical justification is out of the scope of this work. Convolution layer is followed by ReLU layer. There is no change in the dimension of the output of this layer. Next is the Max pooling layer with size 2 and stride 1. Since, our network depth is less so, keeping stride 1 in Max pooling layer was a design choice to avoid loss of information. Total number of learnable parameters were only  ~118,000  as compared to heavier networks like AlexNet (~56,000,000), VGG16 (~134,000,000), VGG19(~139,000,000), ResNet50(~23,000,000), InceptionV3(~22,000,000), and DenseNet121(~7,000,000). 
 After fixing the network architecture, first branch 1 of the network was trained. Trained Branch 1 (CNN layers) were tested on the test set and the results ( OUTPUT -1 ) are accumulated in Table \ref{table6}. Applying transfer learning approach \cite{torrey2009transfer}, we used  MLP for final classification outcome (OUTPUT - 2) instead of 1-of-4 classification output from a softmax layer. Trained CNN backbone has been used as fixed feature extractor. The approach is stage-wise learning and the output 2 observations are recorded in Table \ref{table7} under the column named as "stage-wise".
\subsection{End-to-End Training}
\par We performed end-to-end training of our three layer CNN architecture (CNN\textsubscript{3L} to show the performance of our network when the learning happened from end-to-end rather than stage wise. By end-to-end learning, we mean that both CNN network (CNN\textsubscript{3L}) and MLP network were trained together using the intermediate outputs of CNN\textsubscript{3L} as inputs to the MLP network. So, MLP network received two sets of input features, activated features from the last Max pooling layer of CNN\textsubscript{3L} and the other from the pre-calculated handcrafted features. These two set of features were concatenated before giving them as input to the MLP layers. At the end of the training, two outputs were accumulated, one from the CNN\textsubscript{3L} 1 of 4 softmax outputs and another from the MLP layer classification. The diagram showing the architectural setup is shown in Figure \ref{fig2}.  We performed two experiments in different settings. 
\begin{enumerate}
\item Branch 1 (CNN layers) of the network (refer Figure \ref{fig2}) was trained keeping branch 2 (MLP layers) inactive by turning off the switch and the results of 1-of-4 classification layer of CNN was recorded. 
\item Both branch 1 and branch 2 were trained simultaneously to obtain two outputs, one from CNN layers (OUTPUT-1) and second from MLP layers (OUTPUT-2) (refer Figure \ref{fig2}). The switch was connected in this case.
\end{enumerate}
We have reported the output of MLP classification of combined feature set (OUTPUT-2) in Table \ref{table7}. 

\section{Results and Discussion}

The aim of this work is to highlight the different outcomes of classifying the database with different settings. To represent the outcomes in order of increasing classification performance we have divided our results in the following categories. 
\begin{enumerate}
\item Classification output of only deep learning features from softmax layer of CNN\textsubscript{3L} 
\item Classification results of combined feature set using MLP (O\textsubscript{L}ConvNet) - Stagewise.
\item Classification results of combined feature set using MLP (O\textsubscript{L}ConvNet) - End-to-end. 
\end{enumerate}

\subsection{Results}
Data imbalance has been eliminated using the method described in Section 4.1. We evaluated the proposed model on the basis of F1-scores and multiclass AUC. Multiclass AUC is calculated using the prediction scores given by the softmax function described in (2). 
\par Table \ref{table6} shows the comparative classification performance of MLP and softmax CNN networks, separately. OL features classified the nuclei classes using MLP classifier (Branch - 2 of Fig. \ref{fig2}) while DL features were propagated along the fully connected and classification layers of CNN networks (Branch - 1 of Fig. \ref{fig2}). To visualize the network flow in both OL classification and deep learning classification, connection through the switch in Fig. \ref{fig2} was broken to create two separate branches where, both the branches were trained mutually exclusively.  We compared the OL features performance with CRImage \cite{yuan2012quantitative} (Table \ref{table6}) which also calculates features after the segmentation of the nuclei using basic thresholding, morphological operations, distance transform, and watershed. CRImage also calculates statistical features from the segmented nucleus but lacks visual, shape and texture features. Besides statistical features, incorporation of histopathologically deduced features such as nuclei size, shape, area, color, and texture hold direct relevance to the dataset and therefore the absence of such features prevents CRImage to perform well on the dataset.  So, from the observations reported in Table \ref{table6}, we deduced that our OL features, though only a small number, performed better than   \cite{yuan2012quantitative}. Deep models CNN\textsubscript{3L}, AlexNet, VGG16, VGG19, ResNet50, InceptionV3, and DenseNet121 when tested (without OL features) on the dataset produced F1-score and AUC better than OL features by a huge margin due to exhaustive feature set produced after convolution operations. These features are non-interpretable but have been known to perform quite well in classification tasks. Hence, Table \ref{table6} reports the individual performance of OL features and DL features and proves the statements made in the 'Introduction' Section about the need to shift from traditional feature engineering to convolutional feature learning. 

\begin{table}[htbp]
\renewcommand{\arraystretch}{1.3}
\caption{Comparison between methods stratified by classifier}
\centering
\begin{tabular}{|c|c|c|c|c|c|}
\hline
\multicolumn{2}{|c|}{\textbf{Method}}&\multirow{2}{*}{\textbf{Precision}}&\multirow{2}{*}{\textbf{Recall}}&\multirow{2}{*}{\textbf{F1-Score}}&\multirow{2}{*}{\textbf{Multiclass AUC}}\\
\cline{1-2}
\textbf{Backbone}&\textbf{Classifier}&&&&\\
\hline
\textbf{Only Object Level}&{MLP}&0.5154&0.5156&{0.5135}&{0.7857}\\
\hline
\textbf{CRImage\cite{yuan2012quantitative}}&{SVM (RBF kernel)}&*&*&{0.4880}&{0.6840}\\
\hline
\textbf{CNN\textsubscript{3L}}&{Softmax}&0.8043&0.8046&{0.8040}&{0.9441}\\
\hline
\textbf{AlexNet}&{Softmax}&0.8280&0.8281&0.8216&0.9386\\
\hline
\textbf{VGG16}&{Softmax}&0.8693&0.8699&{0.8689}&{0.9757}\\
\hline
\textbf{VGG19}&{Softmax}&0.8575&0.8581&{0.8578}&{0.9701}\\
\hline
\textbf{ResNet50}&{Softmax}&0.8900&0.8893&{0.8892}&{0.9799}\\
\hline
\textbf{InceptionV3}&{Softmax}&0.8164&0.8175&{0.8175}&{0.9538}\\
\hline
\textbf{DenseNet121}&{Softmax}&0.8784&0.8706&{0.8706}&{0.9756}\\
\hline
\end{tabular}
* data unavailable
\label{table6}
\end{table}
\par Next, we trained the network in Fig. \ref{fig2} stage wise. Features from trained deep learning networks from the previous experiment were concatenated with OL handcrafted features. Then, the second stage of the training of concatenated features was done by MLP classifier. The results obtained on test dataset after MLP training was then reported as combined features classification performance. 
End to end training followed stagewise experiments to analyze the effect of the performance of our network when trained in two ways. We have shown the difference through F1-score, multiclass AUC and cross-entropy loss in Table \ref{table7} which reports performance of classification on a 2-fold cross-validation experiment. The obtained results from stage-wise training, as expected, showed improvement in F1-Score and Multiclass AUC in comparison to individual results obtained after classifying with only DL features and only OL features (Table \ref{table6}). While there is only a 2\% increase in F1-score and 1\% increase in AUC score in case of CNN\textsubscript{3L}, a marked improvement has been recorded in case of deeper pre-trained architectures. Whereas, in the case of end-2-end training, no improvement in the performance metrics has been observed in any of the backbone networks. Also, the higher loss value recorded in the case of end-2-end approach reflects decreased performance in comparison to the stage-wise approach. These cross-entropy loss values highlight that the joint loss propagation after the shared layer (concatenation layer) might have affected the overall performance of the model. The additional OL feature set concatenated simultaneously during training on the shared concatenation layer did not improve the discriminative property of the DL features and hence, the subsequent MLP layers performed in the same way as the fully connected and softmax layers of DL models.  

\par
\begin{figure*} 
\centering
 \subfloat[short for lof][CNN\textsubscript{3L} Test ROC]{
   \includegraphics[width=0.45\textwidth, height=1.5in]{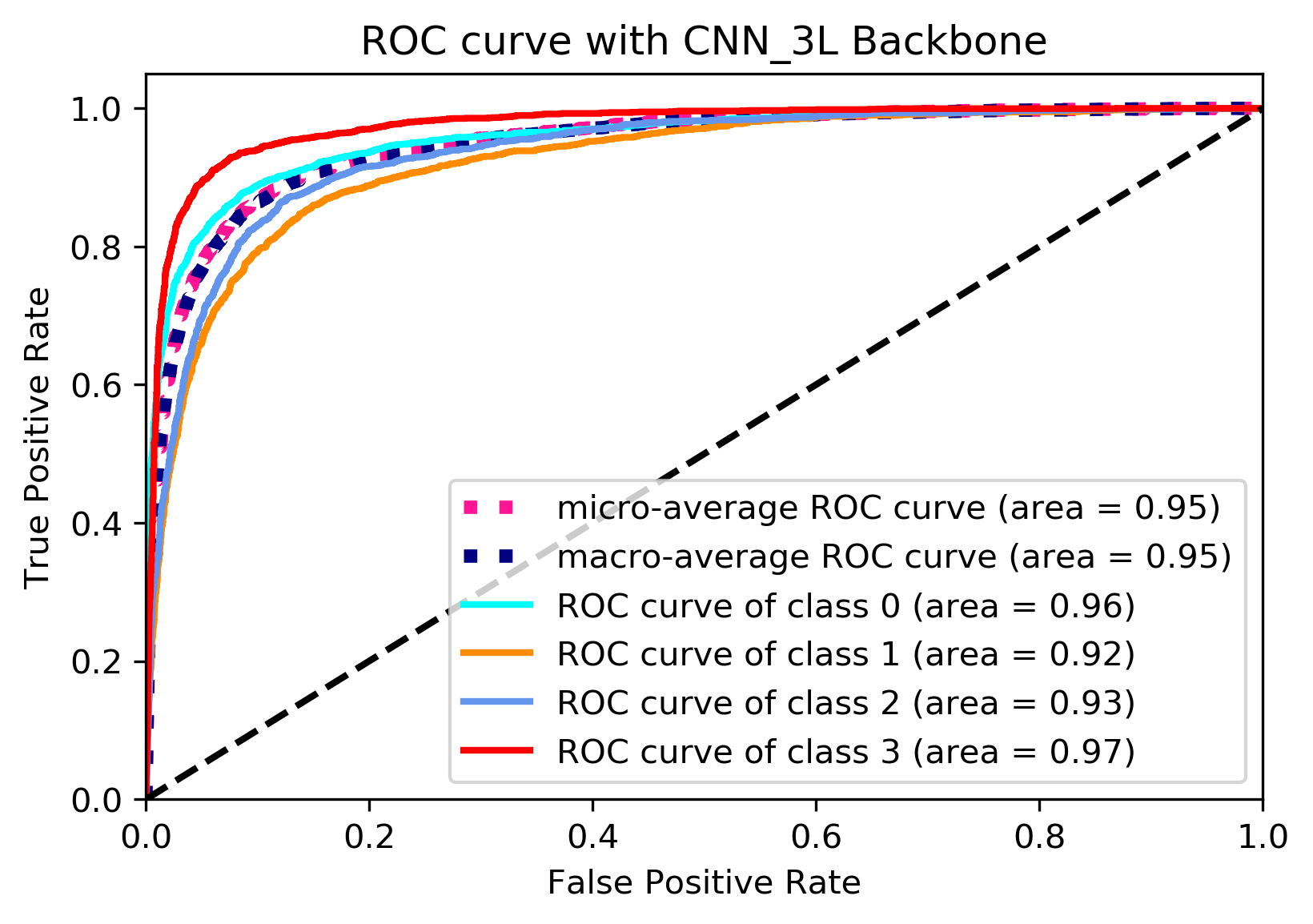}
   \label{subfig:fig1}
 }
 \subfloat[short for lof][AlexNet Test ROC]{
   \includegraphics[width=0.45\textwidth, height=1.5in]{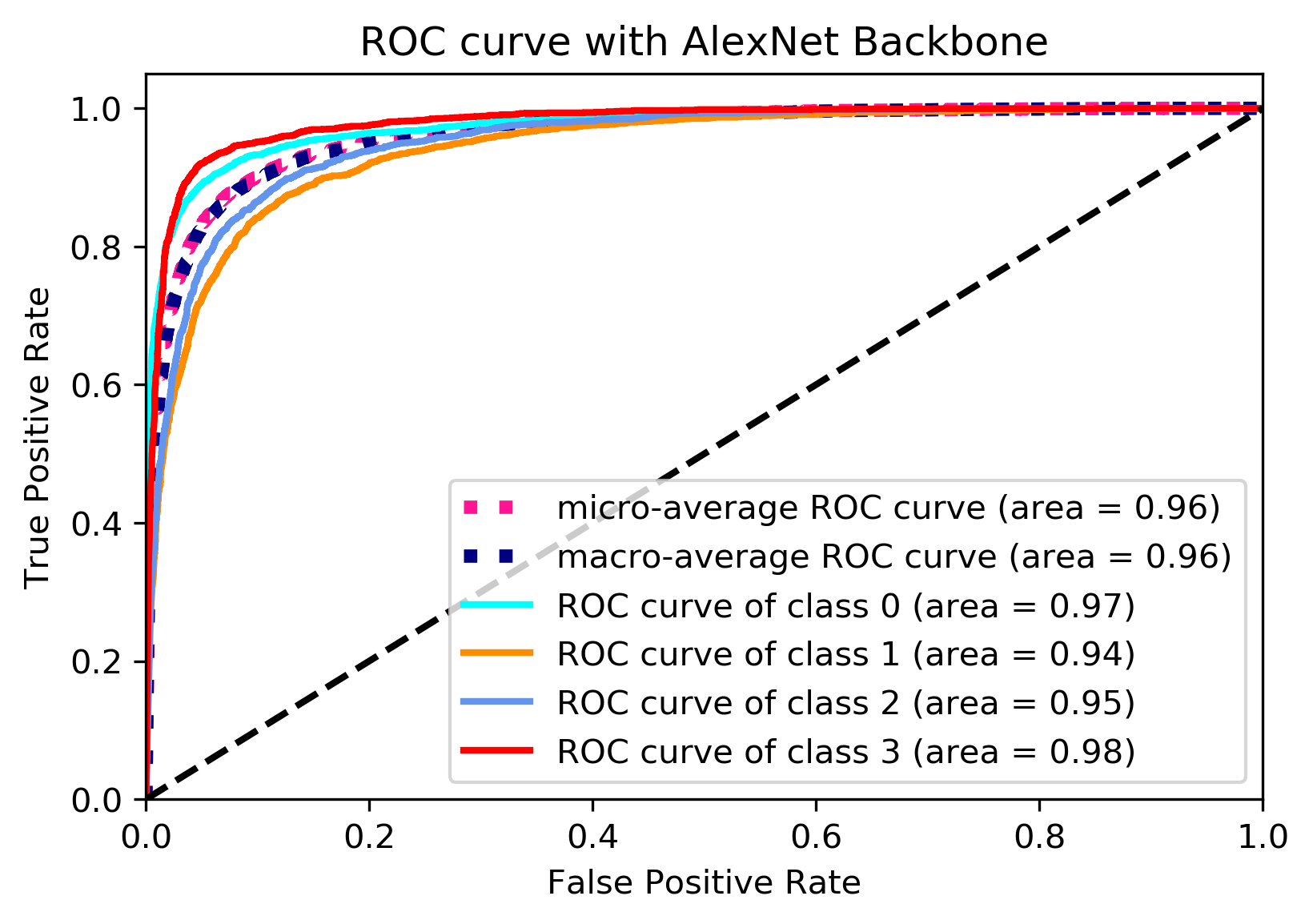}
   \label{subfig:fig2}
}\\
\subfloat[short for lof][VGG16 Test ROC]{
   \includegraphics[width=0.45\textwidth, height=1.5in]{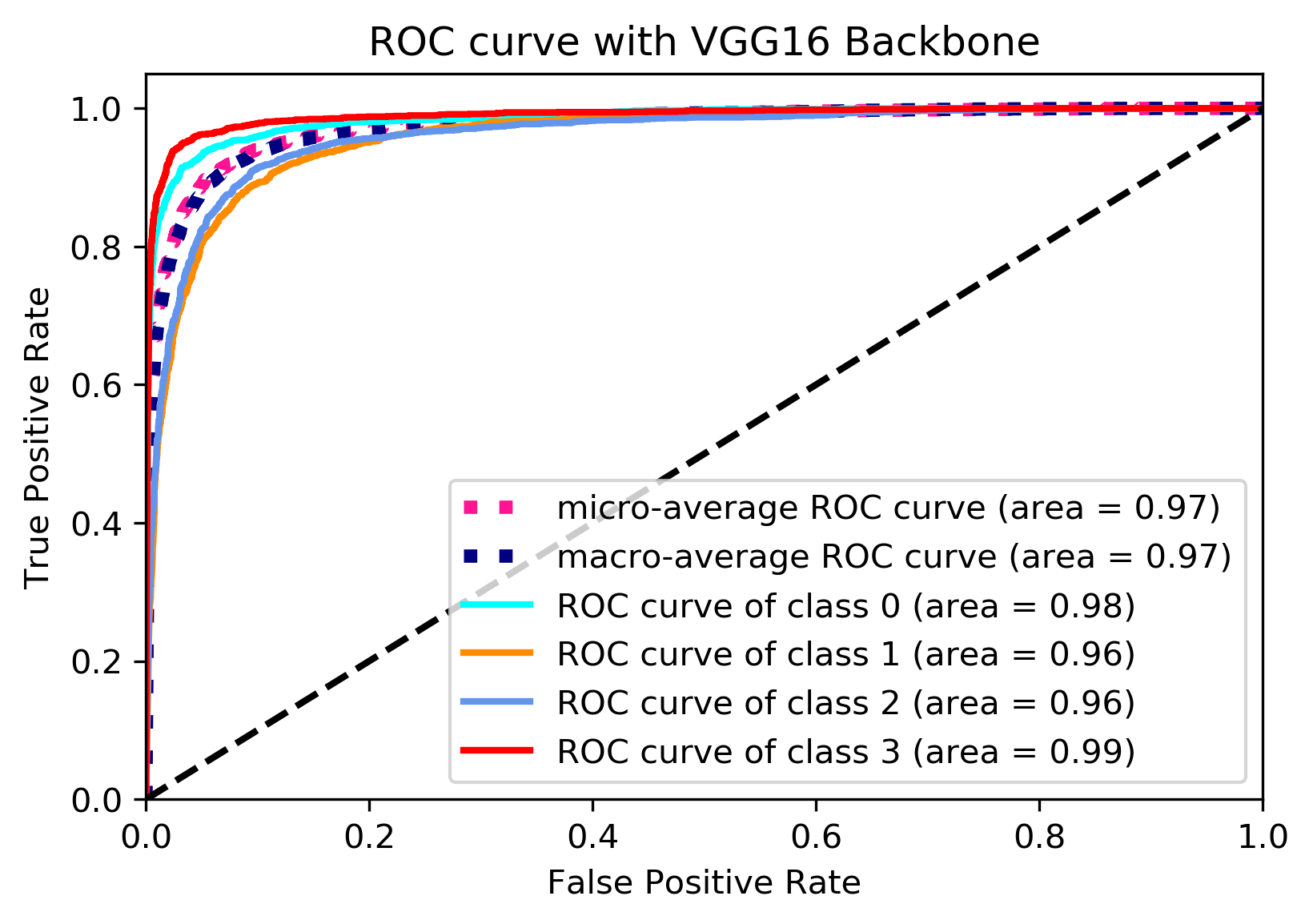}
   \label{subfig:fig3}
 }
 \subfloat[short for lof][VGG19 Test ROC]{
   \includegraphics[width=0.45\textwidth, height=1.5in]{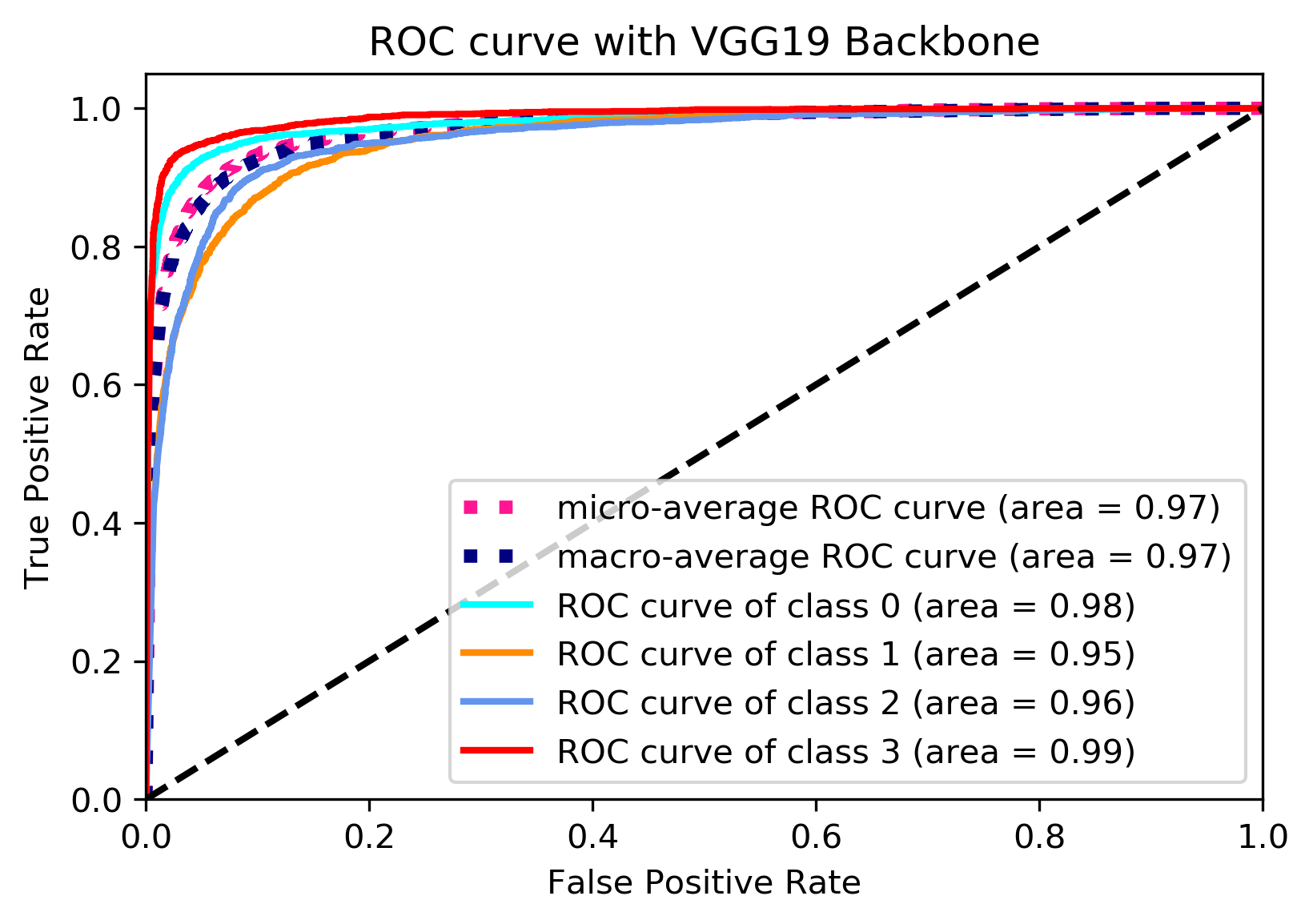}
   \label{subfig:fig4}
}\\
 \subfloat[short for lof][ResNet50 Test ROC]{
   \includegraphics[width=0.45\textwidth, height=1.5in]{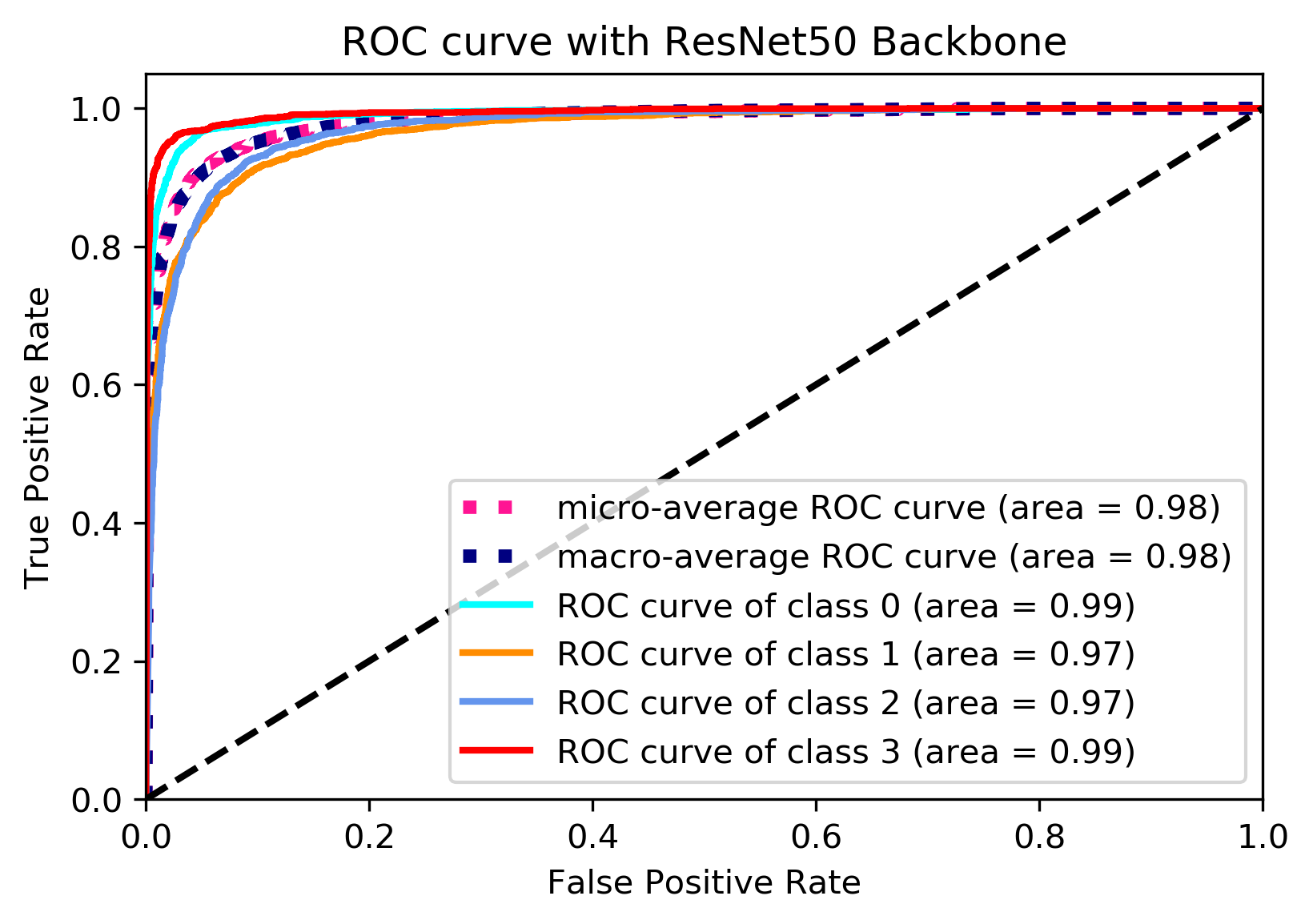}
   \label{subfig:fig5}
}
 \subfloat[short for lof][InceptionV3 Test ROC]{
   \includegraphics[width=0.45\textwidth, height=1.5in]{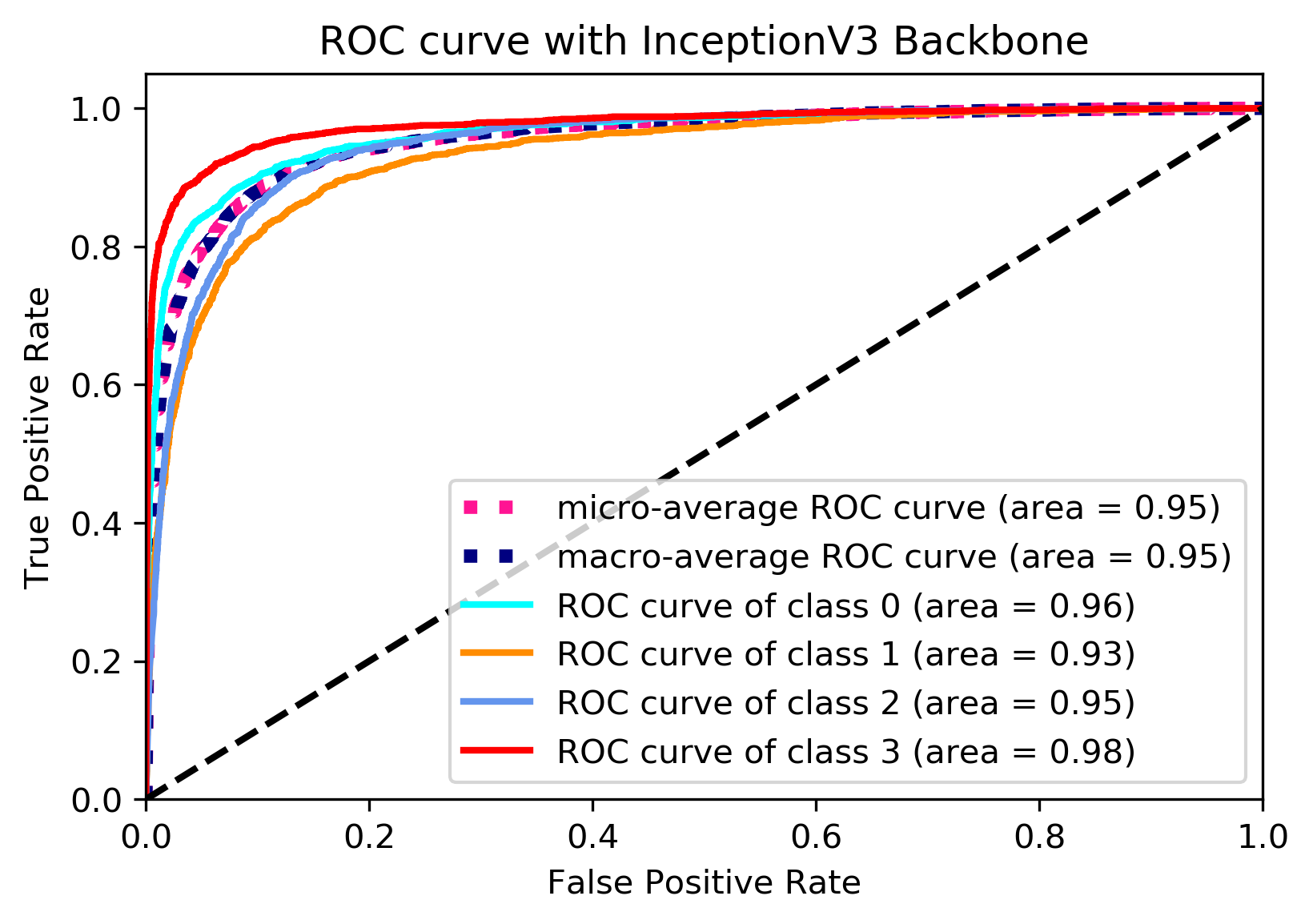}
   \label{subfig:fig6}
}\\
 \subfloat[short for lof][DenseNet121 Test ROC]{
   \includegraphics[width=0.45\textwidth, height=1.5in]{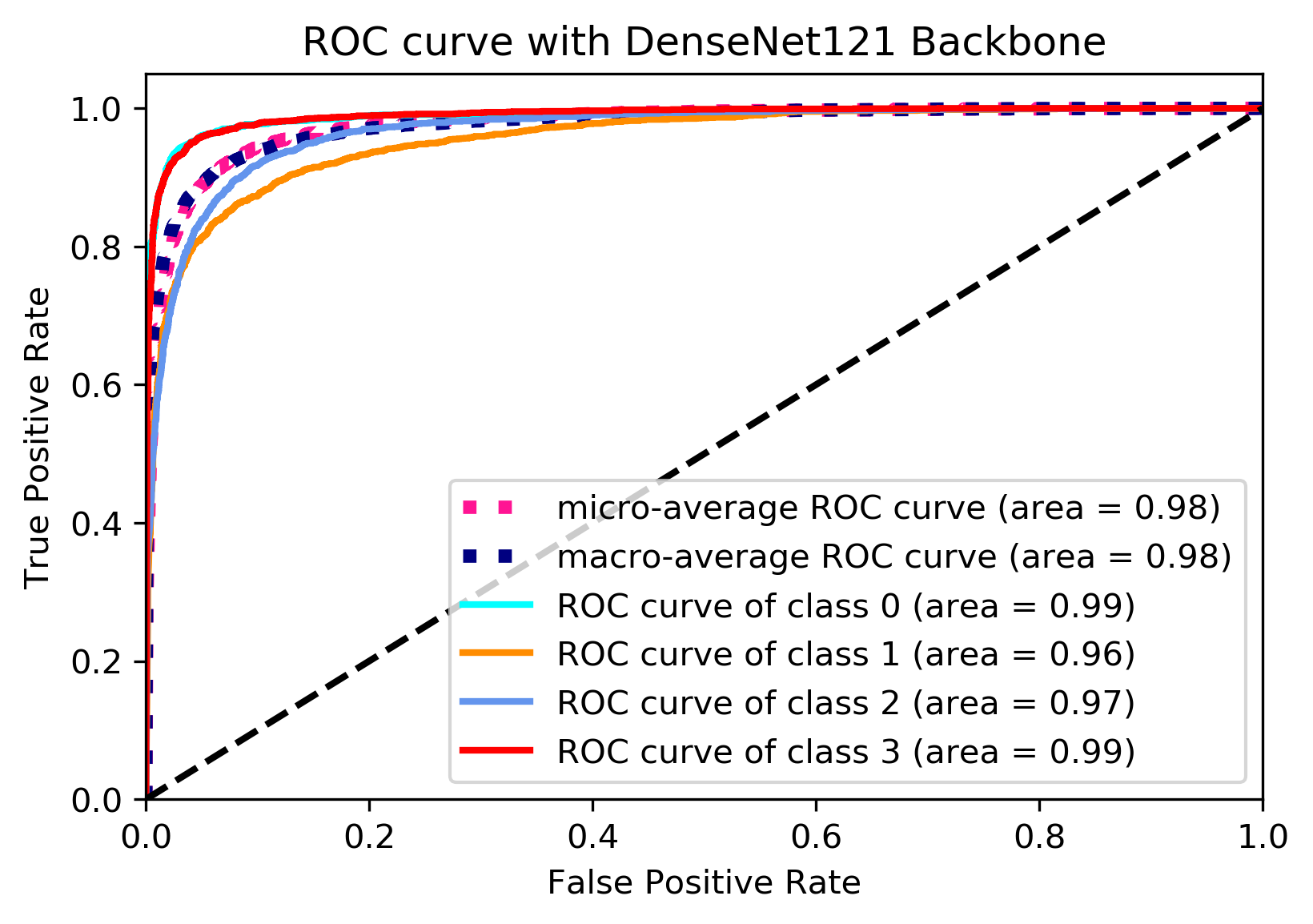}
   \label{subfig:fig7}
}
\caption[short for lof]{ROC curves from four backbone networks using O\textsubscript{L}ConvNet (end-2-end learning)}
\label{fig6}
\end{figure*}

 Fig. \ref{fig6} show the ROC curves obtained after end-2-end training on all four networks. The subfigures \ref{subfig:fig1}, \ref{subfig:fig2}, \ref{subfig:fig3}, \ref{subfig:fig4}, \ref{subfig:fig5},   \ref{subfig:fig6}, and \ref{subfig:fig7} shows the ROC curves of all four classes with respect to seven backbone networks, discriminated through four colors. The figure also mentions the AUC value for individual classes. Micro and Macro average of four ROC curves from each class show similar value because our dataset is balanced. The micro-average method calculates the sum of true positives, false positives, and false negatives for different sets whereas, in Macro-average method, the average of the precision and recall is taken into account. Micro Average is preferred when there is a class imbalance problem. The Figure also labels AUC values for each class. After recording the values we observed a dip of 2\% to 4\% for class 2 (Fibroblast) and 3 (Inflammatory), whereas, the AUC values for classes 1 (Epithelial) and 4 (Miscellaneous) are comparable across all four networks. The decrease in performance can be attributed to the number of samples which are relatively less for class 2 (7275) and class 3 (6970) than class 1 (7722) and class 4 (7804). Also, in the case of Fibroblasts (class 2), the long spindle-shaped cell barely has a visible nucleus which prevents segmentation algorithms to detect the nucleus area effectively. Consistent class-wise performance across all four backbones and high AUC are some of the other observations deduced from Figure \ref{fig6}. 

\par
\begin{figure}[htbp]
\centering
\includegraphics[width=\textwidth, height=3in]{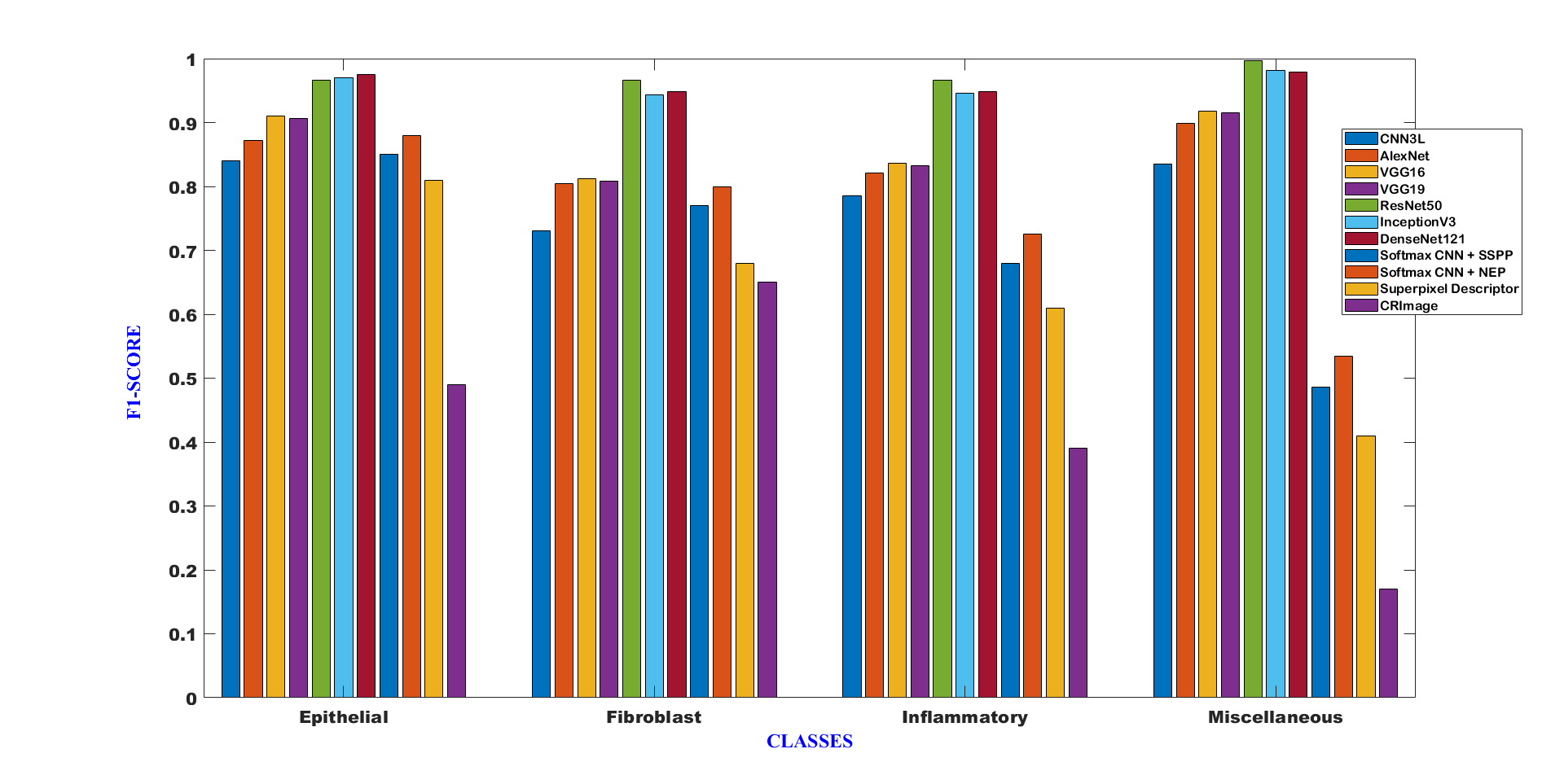}
\caption{Comparative results for nucleus classification stratified with respect to class label }
\centering
\label{fig7}
\end{figure}

 2-fold cross validation performance metrics with CNN\textsubscript{3L} along with deep architectures AlexNet\textsubscript{H}, VGG16\textsubscript{H}, VGG19\textsubscript{H}, ResNet50\textsubscript{H}, InceptionV3\textsubscript{H},and DenseNet121\textsubscript{H}, where H stands for Hybrid, are reported in Table \ref{table9} and Figure \ref{fig7}.  The O\textsubscript{L}ConvNet metrics recorded in Table \ref{table9} are stagewise observations from OUTPUT - 2 of the network obtained after combined feature set testing. We compared our results with Softmax CNN + SSPP (Standard Single Patch Predictor) and Softmax CNN + NEP (Neighboring Ensemble Predictor)  \cite{sirinukunwattana2016locality} architectures used for classification of nuclei from the nuclei dataset cited by \cite{sirinukunwattana2016locality}. The authors in this article worked on the theory that the pixel of interest is likely the center of the nucleus and using this theory they formulated the classification algorithm by spatially constraining the high probability pixels in the vicinity of the centers of nuclei. They proposed two algorithms called Neighboring Ensemble Predictor (NEP) and  Standard Single Patch Predictor(SSPP) which when coupled with SC-CNN (Spatially Constrained- CNN) produced the classification results as mentioned in Table \ref{table7}. The drawback of their method was in building a complex target specific model which lead to low classification performance. Their CNN architecture is custom but they didn't test their methodology on deeper pre-trained architectures which might have performed better with or without their elaborate model. Further comparison with superpixel descriptor \cite{sirinukunwattana2015novel} and CRImage on the same dataset show that both the methods performed relatively poor and that our method exhibited a higher performance regardless of the backbone architecture used. The motivation behind devising Superpixel descriptor was to distinguish the area with the different histologic pattern. This descriptor lacks direct features related to the visual appearance of the nucleus, like color, texture, and shape, thus yielded lower classification performance. Whereas, CRImage \cite{cirecsan2013mitosis} only calculates segmented nuclei features. Segmented nuclei features are insufficient to classify complex nuclei patterns due to either weak staining or presence of overlapping boundaries. This prevents CRImage to perform well in this dataset. We have shown the same through our only OL features classification results in Table \ref{table6}. Figure \ref{fig7} reports comparative class-wise classification performance of all the methods. F1-score of Miscellaneous class show a marked jump in values obtained from O\textsubscript{L}ConvNet and an improvement of more than ~47\% is recorded between the highest and lowest performing algorithms which are ResNet50 and CRImage, respectively. For Epithelial and Fibroblast, the class wise performance of $CNN_{3L}$ was marginally lesser than SSPP but, a difference of 4\% to 7\% was recorded with NEP whereas, in case of DL models used in O\textsubscript{L}ConvNet, the F1-scores have been consistently better than all the algorithms used for comparison. Same can be interpreted from Figure \ref{fig7} where the classwise performance of O\textsubscript{L}ConvNet backbones is consistently better than contemporary algorithms.  
\begin{table*}
\renewcommand{\arraystretch}{1.3}
\caption{Performance parameters from stage wise and end-2-end learning experiments (OUTPUT - 2)}
\label{table7}
\centering
\begin{tabular}{|c|c|c|c|c|c|c|c|}
\hline
\multirow{2}{*}{\textbf{Method}}&\multirow{2}{*}{\textbf{Backbone}}&\multicolumn{3}{|c|}{\textbf{Stagewise}}&\multicolumn{3}{|c|}{\textbf{End-2-End}} \\
\cline{3-8}
&&\textbf{F1-score}&\textbf{AUC}&\textbf{loss}&\textbf{F1-score}&\textbf{AUC}&\textbf{loss}\\
\hline
\multirow{7}{*}{\textbf{O\textsubscript{L}ConvNet}}&\multirow{1}{*}{CNN\textsubscript{3L}}&{0.8243}&{0.9587}&{0.1211}&{0.8084}&{0.9500}&{1.2750}\\
\cline{2-8}
&\multirow{1}{*}{AlexNet\textsubscript{H}}&{0.9542}&{0.9903}&{0.0915}&{0.8359}&{0.9600}&{1.0860}\\
\cline{2-8}
&\multirow{1}{*}{VGG16\textsubscript{H}}&{0.9569}&{0.9923}&{0.1036}&{0.8731}&{0.9700}&{0.9407}\\
\cline{2-8}
&\multirow{1}{*}{VGG19\textsubscript{H}}&{0.9546}&{0.9879}&{0.1221}&{0.8675}&{0.9700}&{1.1213}\\
\cline{2-8}
&\multirow{1}{*}{ResNet50\textsubscript{H}}&{0.9677}&{0.9973}&{0.0272}&{0.8892}&{0.9799}&{0.7335}\\
\cline{2-8}
&\multirow{1}{*}{InceptionV3\textsubscript{H}}&{0.9618}&{0.9963}&{0.0309}&{0.8175}&{0.9538}&{1.0795}\\
\cline{2-8}
&\multirow{1}{*}{DenseNet121\textsubscript{H}}&{0.9616}&{0.9961}&{0.0318}&{0.8706}&{0.9756}&{0.7972}\\
\hline
\end{tabular}
\end{table*}

\begin{table}[htbp]
\renewcommand{\arraystretch}{1.3}
\caption{Comparative Performance parameters for Nucleus Classification}
\label{table9}
\centering
\begin{tabular}{|c|c|c|c|c|c|}
\hline
\textbf{Method}&\textbf{Backbone}&\textbf{Precision}&\textbf{Recall}&\textbf{F1-Score}&\textbf{Multiclass AUC} \\
\hline
\textbf{Softmax CNN + SSPP \cite{sirinukunwattana2016locality}}&{CNN}&*&*&{0.7480}&{0.8930}\\
\hline
\textbf{Softmax CNN + NEP \cite{sirinukunwattana2016locality}}&{CNN}&*&*&{0.7840}&{0.9170}\\
\hline
\textbf{Superpixel Descriptor \cite{sirinukunwattana2015novel}}&{-}&*&*&{0.6870}&{0.8530}\\
\hline
\textbf{CRImage \cite{yuan2012quantitative}}&{-}&*&*&{0.4880}&{0.6840}\\
\hline
\multirow{7}{*}{\textbf{O\textsubscript{L}ConvNet}}&\textbf{CNN\textsubscript{3L}}&0.8241&0.8245&{{0.8243}}&{{0.9587}}\\
\cline{2-6}
&\multirow{1}{*}{\textbf{AlexNet\textsubscript{H}}}&0.9578&0.9577&0.9578&{0.9953}\\
\cline{2-6}
&\multirow{1}{*}{\textbf{VGG16\textsubscript{H}}}&0.9611&0.9610&{0.9610}&{0.9960}\\
\cline{2-6}
&\multirow{1}{*}{\textbf{VGG19\textsubscript{H}}}&0.9544&0.9548&{{0.9546}}&{{0.9879}}\\
\cline{2-6}
&\multirow{1}{*}{\textbf{ResNet50\textsubscript{H}}}&\textbf{0.9676}&\textbf{0.9678}&{\textbf{0.9677}}&{\textbf{0.9973}}\\
\cline{2-6}
&\multirow{1}{*}{\textbf{InceptionV3\textsubscript{H}}}&0.9618&0.9618&{{0.9618}}&{{0.9963}}\\
\cline{2-6}
&\multirow{1}{*}{\textbf{DenseNet\textsubscript{H}}}&0.9616&0.9617&{{0.9616}}&{{0.9961}}\\
\hline
\end{tabular}
* data unavailable
\end{table}

\subsection{Observations and Discussions}
\par The experiments performed in this study gave some interesting points for discussion. Formally, whenever we performed deep learning based classification tasks, our focus generally remained on improving the classification performance of the architecture by fine-tuning or transfer learning. Transfer learning is generally used in the case when there are fewer samples of similar data. In our case as well, samples were not enough for deep architectures to generalize well on the dataset. Moreover, pre-trained weights of discriminative nuclei types to transfer learn on our database was also not available. Also, the intention to not use pre-trained publically available ImageNet weights was because our dataset is very dissimilar to ImageNet. Hence, we chose to build our own CNN layers (CNN\textsubscript{3L}) and fine-tune them to produce optimized performance parameters. 

\par DL features from CNN\textsubscript{3L} and OL features representing visual and structural properties of nuclei were concatenated to produce a combined feature set which consequently improved classification results. F1-Score and multiclass AUC values recorded in Table \ref{table6} reflected the individual performance of the traditional handcrafted OL features with MLP classifier and DL features with Softmax classifier whereas, F1-Score and multiclass AUC values in Table \ref{table9} show the performance of the O\textsubscript{L}ConvNet network with combined feature sets. These observations reflect that even if the OL feature set was very small (total 9) relative to CNN\textsubscript{3L} feature length (48,400) , a marked difference between 1-of-4 softmax classification (OUTPUT 1 of Fig. \ref{fig2}) and O\textsubscript{L}ConvNet classification (OUTPUT 2 of Fig. \ref{fig2}) reflect the applicability and importance of including object-specific features. Further, the high performance metric values in Table \ref{table9}, exhibited by deeper pre-trained architectures ($AlexNet_{H}$, $VGG16_{H}$, $ VGG19_{H}$, $ ResNet50_{H}$, $ InceptionV3_{H}$, $ DenseNet121_{H}$) shows  that increasing depth of the network although increase the capability to extract discriminative DL features and hence better F1-score and AUC from $CNN_{3L}$ (as reflected in Table \ref{fig6}) but, the discriminative ability of the DL feature set is enhanced many folds when combined with OL features. This statement can be verified from Table \ref{table7} which shows the increase of ~10\% in F1-score and AUC values of these pre-trained architectures.  However, training O\textsubscript{L}ConvNet jointly for classification (end-2-end ) ( Table \ref{table7}) using a combined feature set did not affect F1-score and AUC values. This is likely because we trained the network with the combined loss of CNN classification layer and MLP output layer. The combined cross-entropy loss affected the performance of the joint network. Therefore, from Table \ref{table7} we observe that the loss values are higher in end-2-end training in comparison to stage-wise training. Hence, stagewise 2-fold cross-validation results are preferred over end-2-end classification. The results were also suggestive of the fact that the classification performance has most certainly increased quite dramatically as the levels of convolution layers increased from just three layers in CNN\textsubscript{3L} to 121 layers in DenseNet121. In all the cases, though, our hypothesis remained true and our architectural setup provided the room to make modifications in the configuration of backbone networks.

 
\par The question that can be raised at this point is why to use CNN\textsubscript{3L} when we are getting better performance with deeper architectures. In support of CNN\textsubscript{3L}, when we look in Table \ref{table7} and observe cross-entropy loss of all backbones in stagewise learning, we see that CNN\textsubscript{3L}  has comparable loss value with deeper networks but, the number of parameters and the training time taken by AlexNet, VGG16, VGG19, ResNet50, InceptionV3, and DenseNet121 shown in Table \ref{table8} is much larger than CNN\textsubscript{3L}. This opens up the discussion for the need of using an optimal number of layers to achieve satisfactory performance by an application instead of very deep CNN models. To further strengthen our point, we know that in deep learning, a number of successful early and recent state of the art architectures such as AlexNet, VGG16, VGG19, , ResNet50, InceptionV3, and DenseNet121 etc., were developed to increase the abstractness in extracting deep and better quality feature maps, primarily for the purpose of classification. However, the standard datasets used to test these networks were real life natural images. 

\par Medical data is a completely different modality that has high variations in features from the same class. Therefore, the state of the art deep network models on such datasets do not perform well and suffer from high validation and testing loss. Another main reason for their failure is a scarcity of labeled data volume in the medical image domain.  To cope up with these limitations, every new model that is being developed for medical image data classification is function specific and does not address the global problems in the domain. Most of the recent literature in the classification of structures in histology images do not use raw images for deep learning models. A fair amount of pre and post processing of data is required to enhance the performance which consequently, hampers the generalization capability of the model. So, instead of building novel models with less or limited applicability in the cell nuclei image classification problem, it seemed better to change the way these models are used. With simple architectural change and introduction of visually and structurally relevant handcrafted feature set, through experiments, we have established gain in performance values. Our model is flexible in a way that any deep model can be fitted in the architecture to extract deep features. Another highlight of our work is to show how the addition of a handful of basic OL handcrafted features can bring a notable change in the final classification output. We do not need to design special descriptors or use complex handcrafted features to complement deep learning features. Instead, a small number of discriminative OL features in case of nuclei classification like color, texture, and shape can enhance the discriminative capability of the neural network classifier. 

\par While we agree that the architecture is simple, we first need a fine-tuned and flexible model that scales well with the number of options of novel models available today. More than that, we need a compact model with less training time ( a comparatively shallow convolutional network ) that fairly works well and give comparable results with deeper architectures.  We compared the time taken by each of the methods to train in Table \ref{table8}. We observed that CNN\textsubscript{3L} took the least amount of time and very huge differences were noted when compared with ResNet50, InceptionV3, and DenseNet121. Time may be regarded as an insignificant paradigm with advanced computer systems available today. However here, it is important to mention that, we trained our algorithms using three NVIDIA GeForce GTX 1080 Ti GPUs with 11GB RAM each, in parallel. This much computing power is not still really common in most of the places and, also it becomes difficult to train heavy deep networks in light mobile applications or hand-held systems. These much computing resources are impossible to accommodate in lighter applications as of yet and hence, shallow networks that work fairly well for general diagnostic procedures can help in reducing space and time constraints for such systems. In this research,  we are dealing with time sensitive application where the quick delivery of classification results are important to help pathologists to proceed for further analysis in cancerous tissues and diagnose cancer as soon as possible. Besides time, we would also like to highlight that our architecture CNN\textsubscript{3L} is not using pre-trained weights of ImageNet which has several other advantages such as, applications can use the custom size of their dataset directly without the need of resizing it to conform according to the size of ImageNet data. This is beneficial at times where very large or very small images may lose quite a significant amount of details when scaled. While one may argue in this case that we could have trained all DL backbones from scratch without the need of using pre-trained weights. However, these deep networks require a certain minimum size of the image (48 x 48 in case of VGG16 and VGG19, 197 for resNet50 and 221 for DenseNet121) at the input layer to train.  Hence, the limitation of these state of the art deep learning architectures with complex datasets is too much to ignore. Additionaly, for training such deep networks from scratch, we would need a very large number of dataset samples for the networks to learn efficiently.
Table \ref{table8} summarizes the time is taken and the number of trainable parameters used by CNN\textsubscript{3L}, AlexNet, VGG16, VGG19, ResNet50, InceptionV3, and DenseNet121 to train the dataset.
\begin{table}[htbp]
\renewcommand{\arraystretch}{1.3}
\caption{Time and trainable parameters comparison between backbone architectures}
\label{table8}
\centering
\begin{tabular}{|c|c|c|}
\hline
\textbf{Method}&\textbf{No. of parameters}&\textbf{Time (seconds)}\\
\hline
\textbf{CNN\textsubscript{3L}}&\textbf{118 thousands}&\textbf{303}\\
\hline
\textbf{AlexNet}&{56 millions}&{3304}\\
\hline
\textbf{VGG16}&{134 millions}&{1408}\\
\hline
\textbf{VGG19}&{139 millions}&{1420}\\
\hline
\textbf{ResNet50}&{23 millions}&{18951}\\
\hline
\textbf{InceptionV3}&{22 millions}&{13314}\\
\hline
\textbf{DenseNet121}&{7 millions}&{24515}\\
\hline
\end{tabular}
\end{table}
\\
Hence, a simple architectural setup whose components can be modelled as per the dataset requirement and which uses the combination of OL features and shallow, yet incorporating the properties of DL model - CNN\textsubscript{3L} in case of complex dataset such as ours has proved to be a better approach than the traditional OL models or trending DL algorithms, alone that do not allow changes in the architecture and require specific configurations to work well. It is important to note that there are no standard datasets yet in case of histological nuclei images so, comparing various methods mentioned in the literature on only one database does not guarantee to give expected results. According to the free lunch theorem, which is most certainly applicable in case of medical image databases, there is no global algorithm that could be developed to give good results across all kinds of histopathological data. Each experiment is conducted with different dataset acquired by the research team themselves or from the pathologists, whose property changes with the location of the disease. Their results are validated using different performance metrics. Hence, standard datasets and ground truth in the case of complex histopathological cancer images is a current challenge in this field of research. 

\section{Conclusion}
The knowledge about the cell of origin of a tumor may benefit doctors to treat the tumor more effectively since a correct classification will greatly increase the biological understanding of carcinogenesis. Using this motivation in this paper, we have built our classification model by emphasizing on a hybrid and flexible model that can incorporate the two wide domains of features,  the age-old traditional object-level features such as intensity, texture, and shape features and recent deep learning based features. While object-level features have proved their efficiency in various domains and purposes of biomedical image processing,  including cancer-based disease recognition and classification, the current trend has drastically shifted towards using various deep learning methods. 
Our work tried to highlight through the results that using only deep learning might not work in case of all datasets. Therefore, a need to develop a shallow yet effective architecture such as CNN\textsubscript{3L} incorporated in our proposed skeleton called O\textsubscript{L}ConvNet, that could robustly combine the benefits of object level features in this study, motivated our work. Moreover, to guarantee a wide range of applicability, a model that is easy to understand, and deploy is required, also, which has the flexibility to incorporate both deeper, more efficient networks like AlexNet, VGG16, VGG19, ResNet50, InceptionV3, and DenseNet121, and shallow, light model like CNN\textsubscript{3L} so that the O\textsubscript{L}ConvNet can be adapted to a wider range of applications. 
The results were encouraging and our network performed better than the recent state of the art implementation on the same dataset. The future work could be to incorporate better algorithms that combine the best of both the worlds i.e. traditional object level and deep learning features. Approaches can differ with better classifiers as well. In conclusion, our method opens the possibility of further research in developing more robust nuclei classification models that can scale well on all kind of datasets.

\section{Acknowledgments}
This research was carried out in Indian Institute of Information Technology, Allahabad and supported by the Ministry of Human Resource and Development, Government of India. We are also grateful to the NVIDIA corporation for supporting our research in this area by granting us TitanX (PASCAL) GPU. 

\bibliographystyle{unsrt}
\bibliography{classification}

\end{document}